\documentclass{INTERSPEECH2023}

 \interspeechcameraready

\title{Zambezi Voice: A Multilingual Speech Corpus for Zambian Languages}
\name{Claytone Sikasote$^1$, Kalinda Siaminwe$^1$, Stanly Mwape$^1$, Bangiwe Zulu$^1$, Mofya Phiri$^1$, \\ Martin Phiri$^1$, David Zulu$^1$, Mayumbo Nyirenda$^1$,  Antonios Anastasopoulos$^2$}
\address{
  $^1$University of Zambia, Zambia\\
  $^2$George Mason University, U.S.A }
\email{claytone.sikasote@cs.unza.zm}

\begin{document}

\maketitle
 
\begin{abstract}
This work introduces Zambezi Voice, an open-source multilingual speech resource for Zambian languages. It contains two collections of datasets: unlabelled audio recordings of radio news and talk shows programs (160 hours) and labelled data (over 80 hours) consisting of read speech recorded from text sourced from publicly available literature books. The dataset is created for speech recognition but can be extended to multilingual speech processing research for both supervised and unsupervised learning approaches. To our knowledge, this is the first multilingual speech dataset created for Zambian languages. We exploit pretraining and cross-lingual transfer learning by finetuning the Wav2Vec2.0 large-scale multilingual pretrained model to build end-to-end (E2E) speech recognition models for our baseline models. The dataset is released publicly under a Creative Commons BY-NC-ND 4.0 license and can be accessed through the project repository. \footnote{\url{https://github.com/unza-speech-lab/zambezi-voice}} 
\end{abstract}
\noindent\textbf{Index Terms}: speech recognition, speech translation, low resource speech processing, Zambian languages

\section{Introduction}
\label{introduction}

The development of speech-based systems like automatic speech recognition (ASR) and speech-to-text translation (STT) systems requires data resources (text and speech) on which models can be trained and tested. In the recent years, the field of natural language processing (NLP) and speech processing (SP) has remarkably progressed thanks to advances in deep learning methods such as self-supervised learning, i.e., pretrained large speech models like XLS-R~\cite{babu22_interspeech}; availability of massive unlabelled and labelled datasets such as LibriSpeech~\cite{Panayotov2015}, Libri-Light~\cite{Kahn2020} and Common Voice~\cite{Ardila2020} inter alia; and the existence of large-scale computational infrastructure~\cite{Hirschberg2015}. However, most of this progress has been demonstrated in just a handful of high-resourced languages like English, French, or Chinese, leaving behind a vast majority of the world's  over 7000 languages, for which the dearth of data resources prohibits the development of functional speech and language-based systems~\cite{Joshi2020}.

In this paper, we present an initial release of a multilingual speech corpora, Zambezi Voice, containing two sets of speech data resources, \textit{labelled} and \textit{unlabelled} speech datasets for some major native languages of Zambia. The languages despite being considered "major", are still severely under-resourced due to the unavailability or lack of data resources that can be used to develop speech-based systems like ASR. The labelled ASR datasets are comprised of over 80 hours in total of read speech data for four languages: Bemba (28 hours), Nyanja (25 hours), Tonga (22 hours), and Lozi (6 hours). The unlabelled audio collection is comprised of radio broadcast-styled speech data - original/unsegmented (over 350 hours) and segmented (over 168 hours) - for the languages: Bemba, Nyanja, Tonga, Lozi, and Lunda. In addition, we conduct baseline experiments by training E2E ASR models using cross-lingual transfer learning to ascertain its potential and usefulness. To our knowledge, this is the first multilingual speech corpus developed for Zambian languages.

\begin{figure}[t]
  \centering
  \includegraphics[width=5.5cm]{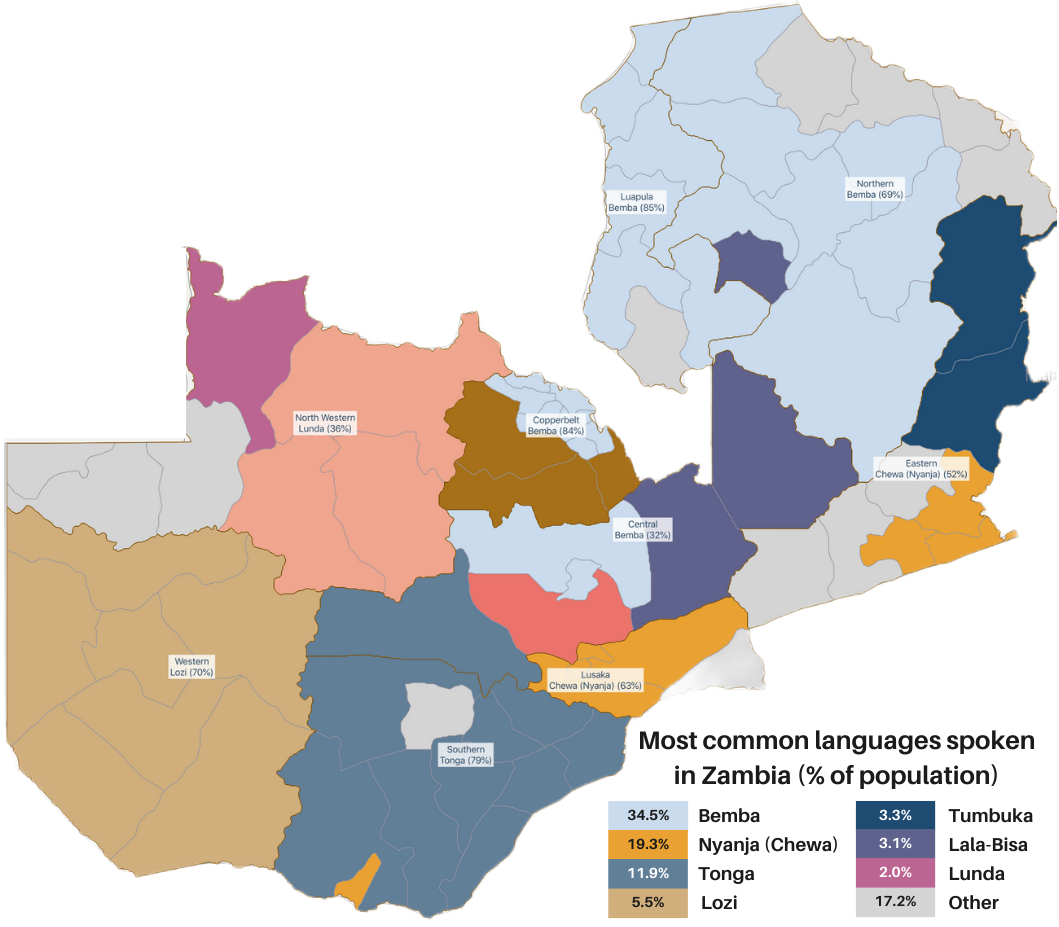}
  \caption{Language map of Zambia showing the most common languages spoken in Zambia (\% of the population) based on the Census Report of 2010\cite{census2010report}. The color codes indicate the primary language spoken in the specific region of Zambia.}
  \label{fig:speech_production}
\end{figure}

\section{Zambian languages}
\label{zambia-society}
Zambia is a diverse and multilingual sub-saharan country endowed, officially, with 73 ethnic/tribal groupings, from which seven spoken language clusters are identified and formed based on factors such as geographical location and mutual intelligibility~\cite{census2010report}. All the languages belong to the Bantu family and their writing systems are based on the Latin script. 
\subsection{Languages}
While this work is ongoing and aimed at curating speech data resources for all the seven official native languages of Zambia, eventually also including other minority languages, in this work the speech data resources we present correspond to only five languages: Bemba, Nyanja, Tonga, Lozi, and Lunda. We briefly introduce these languages below:

\noindent \textbf{Bemba (bem)} \ 
Bemba, also known as IciBemba, is the most widely spoken language in Zambia with an estimation of over 35\% speakers of the entire country population~\cite{census2010report}. It is native to the people of Northern, Luapula, and Muchinga provinces but also spoken in other parts of the country like the Copperbelt (83.9\%), Central (31.8\%), and Lusaka (17.6\%) provinces~\cite{census2010report}. Bemba is also spoken in other countries like the Democratic Republic of Congo (DRC) and Tanzania. 

\noindent \textbf{Nyanja (nya)}  \ 
Nyanja, also known as Chinyanja or Chewa, is principally spoken by people of Eastern and Lusaka provinces\cite{census2010report}. It is the second most widely used language of communication in Zambia with a speaker estimation of 14.8\% of the total country population. Nyanja is also spoken in Malawi where it is referred to as Chewa and used as an official language of communication. 

\noindent \textbf{Tonga (toi)} \ 
Tonga, sometimes referred to as Chitonga, is native to the people of the Southern province, who are also referred to as Batonga. It is the third most spoken language (11.4\%) in Zambia. By province, Tonga is spoken in Southern (74.7\%), central (15\%), and Lusaka (4.3\%) provinces. It has several dialects: Plateau Tonga, Valley Tonga, Leya, Mala, Ndabwe, and Dombe. It is also spoken in parts of other countries like Zimbabwe and Mozambique~\cite{census2010report}.

\noindent \textbf{Lozi (loz)} \ 
Lozi, also sometimes referred to as SiLozi, is native to the people of the Western province(69\%). It is also spoken in other provinces like Southern (4.0\%), Lusaka (1.3\%), and central (1.0\%) provinces~\cite{census2010report}.

\noindent \textbf{Lunda (lun)} \
Chilunda, sometimes referred to as Lunda, is the most populous language in North-Western province of Zambia with an estimation of over 33.8\% speakers of the entire province~\cite{census2010report}. It is also spoken in parts of Angola and the Democratic Republic of Congo. 

\section{Related works}
\label{related-works}

\subsection{Bemba Corpora}

In relation to available speech data resources for Zambian languages, to our knowledge, and at the time of authoring this paper, there are only two speech datasets publicly: BembaSpeech\cite{sikasote-anastasopoulos:2022:LREC}, which provides over 24 hours of read-styled ASR data for Bemba, and BIG-C\cite{sikasote2023big}, a large-scale multimodal and multi-purpose dataset comprised of multi-turn dialogues between Bemba speakers grounded on images, having over 92,000 utterances, amounting to over 180 hours of speech data with corresponding transcriptions and English translations. While these resources are crucial for the development of speech-related systems for Bemba, there exist no similar speech data resources for other languages - over 70 languages. Therefore, this work seeks to address this gap by creating a multilingual speech dataset for Zambian languages.

\subsection{ASR datasets from Radio Broadcasts} 
To mitigate the unavailability of speech recognition datasets, some researchers have used radio broadcasts to build systems. Notable examples are the West African Radio Corpus~\cite{doumbouya2021usingradio}, a multilingual speech dataset containing 142 hours for more than 10 West African languages; Makerere Radio Speech Corpus~\cite{mukiibi-etal-2022-makerere} containing 155 hours of transcribed radio broadcasts that were used to build viable ASR models for Luganda in Uganda; Congolese Speech Radio Corpus~\cite{kimanuka2023speech} containing 741 hours of unlabelled audio collections for 4 major languages  spoken in the Democratic Republic of Congo i.e., Lingala, Tshiluba, Kikongo, and Congolese Swahili. Similarly, \cite{zanon-boito-etal-2022-speech} created speech resources from radio broadcast recordings containing 671 hours of unlabelled audio data in five languages, namely French, Fulfulde, Hausa, Tamasheq, and Zarma, and a 17 hours parallel speech translation corpus to translate Tamasheq spoken utterances to French. These works demonstrate how radio-styled speech data can be used for ASR systems, especially in low-resource environments where data resources are not available or exist in limited form. We believe that our unannotated radio-styled audio collections will prove useful towards the development of speech recognition systems for Zambian languages, especially by leveraging unsupervised or self-supervised learning approaches.

\section{Zambezi Voice}
\label{zambezi-voice}
\subsection{Description}
Zambezi Voice is an open-source multilingual speech corpus containing two collections of speech data resources; labelled and unlabelled datasets. The labelled datasets consist of speech datasets for four (4) languages; Bemba, Nyanja, Tonga and Lozi whereas the unlabelled data collection consists of unannotated TV/Radio broadcast-styled audio collections for Bemba, Nyanja, Tonga, Lozi and Lunda. In this section, we give details of Zambezi Voice. We begin with describing labelled datasets in~\S~\ref{sec:annotated-datasets} followed by unlabelled datasets in~\S\ref{sec:unannotated-datasets}.

\subsection{Labelled datasets}
\label{sec:annotated-datasets}
This initial release contains three labelled speech corpora for Nyanja, Tonga, and Lozi. The datasets consist of read speech recorded from text sourced from publicly available literature books for Tonga and Lozi, and Nyanja translations of a portion of English captions of images from Flickr30K~\cite{plummer2015flickr30k} dataset. Similar to~\cite{sikasote-anastasopoulos:2022:LREC}, we tokenized the text at the sentence level and recorded the sentences using the \textit{elicitation by text} mode of the LIG-AIKUMA mobile application~\cite{gauthier2016lig}. All the speakers recorded in uncontrolled environments, thus some recordings may have background noise. Table~\ref{labelled-datasets} summarises the dataset statistics. All audio files are encoded in Waveform Audio File Format (WAVE) with a single track (mono) and recording with a sample rate of 16kHz.

\begin{table}[th]
	\caption{\label{labelled-datasets} Basic statistics of the labelled datasets.}
	\centering
	\begin{tabular}{lccccc}
		\toprule 
		\textbf{Datasets} & \textbf{} & \textbf{Hours} & \textbf{Utterances} & \textbf{Speakers} & \textbf{}  \\
		\midrule
		\multicolumn{6}{l}{\textit{Bemba (bem):}}\\
		\cmidrule(l){1-2}
		train &  & 22.0 & 12,421 & 9  \\
		dev & & 2.5 & 1,700 & 8  \\
		test & & 2.5 & 1,368 & 3  \\
		\midrule
		total & & 28.0 & 15,489 & 20  \\
		\midrule
		\multicolumn{6}{l}{\textit{Nyanja (nya):}}\\
		\cmidrule(l){1-2}
		train &  & 22.0 & 8,117 & 6  \\
		dev & & 2.0 & 622 & 4  \\
		test & & 1.0 & 428 & 2  \\
		\midrule
		total & & 25.0 & 9167 & 12  \\
		\midrule
		\multicolumn{6}{l}{\textit{Tonga (toi):}}\\
		\cmidrule(l){1-2} 
		train & & 19.5 & 8,340 & 5 \\
		dev & & 1.5 & 551 & 2  \\
		test & & 1.5 & 453 & 2  \\
		\midrule
		total &  & 22.0 & 9,344 & 9  \\
		\midrule
		\multicolumn{6}{l}{\textit{Lozi (loz):}}\\
		\cmidrule(l){1-2} 
		train & & 4.0 & 1,855 & 3  \\
		dev & & 1.0 & 670 & 2  \\
		test & & 1.0 & 399 & 1 \\
		\midrule
		total &  & 6.0 & 2,924 & 6  \\
		\bottomrule
		
	\end{tabular}
\end{table}

\noindent \textbf{Bemba} \  
For the Bemba dataset, we extend prior work~\cite{sikasote-anastasopoulos:2022:LREC} by adding two hours of read speech to the BembaSpeech~\cite{sikasote-anastasopoulos:2022:LREC} ASR dataset. The dataset contains 28 hours recorded by 20 speakers all identified to be racially black. The details of the splits can be found in Table~\ref{labelled-datasets}.

\noindent \textbf{Nyanja}  \ 
The Nyanja dataset contains over 25 hours of speech data recorded by 12 speakers, 4 male and 8 female using the text sourced from Nyanja translations of English captions of images from Flickr30K~\cite{plummer2015flickr30k}. We crowd-sourced five qualified High School teachers of the Nyanja language, who also are native speakers of the language to translate a portion of English image captions.  A total of 28,000 sentences were translated to Nyanja, from which approximately over 9,000 sentences are recorded using the LIG-AIKUMA~\cite{gauthier2016lig} mobile application. All speakers/recorders were students of the University of Zambia and identified as black. The dataset can be used for speech recognition and speech/text translation as well as for image-text tasks since the text is aligned to images.

\noindent \textbf{Tonga} \ 
The Tonga dataset is an ASR corpus containing 22 hours of read speech recorded by 9 native Tonga speakers, 4 male and 5 female, using the text we collected from publicly available literature books sourced from an online library, Lubuto Library Partners.\footnote{\url{https://www.lubuto.org/read-zambian-stories}} The text collected was tokenized at sentence-level and recorded using LIG-AIKUMA~\cite{gauthier2016lig}. All annotators i.e., text collectors,  speakers, and validators identified as black, and they were students of the University of Zambia.

\noindent \textbf{Lozi} \ 
Similar to the Tonga ASR dataset, the Silozi dataset contains read speech recorded using the text sourced from the publicly available literature books from Lubuto Library Partners. It contains 6 hours of speech recorded by 4 native language speakers, 4 male and 2 female, all students of the University of Zambia and identifying as black.

\noindent \textbf{Gender distribution} \ 
In our data curation process, we strived for a diverse speaker gender distribution, but nonetheless, there are imbalances (see Table~\ref{tab:gender_distribution}). The Nyanja and Lozi have more utterances labeled with male voices than female voices. The opposite is true for the Tonga dataset; it has more samples labeled with female voices than male ones.

\begin{table}[th]
	\caption{ Number of speaker gender voice distribution among the labelled datasets.}
	\label{tab:gender_distribution}
	\centering
	\begin{tabular}{lccccc}
		\toprule 
		\textbf{Datasets} & \textbf{} & \textbf{Male} & \textbf{Female} &             \textbf{Unspecified} & \textbf{}  \\
		\midrule
            Bemba &  & 10,396 & 5,047 & 46  \\
		Nyanja &  & 7,083 & 1,604 & 349  \\
		Tonga & & 982 & 5,562 & -  \\
		Lozi & & 1,978 & 946 & -  \\
		\bottomrule
		
	\end{tabular}
\end{table}

\noindent \textbf{Preprocessing} \ 
We adopted the data preprocessing of \cite{sikasote-anastasopoulos:2022:LREC} to ensure accurate and quality content in all the datasets. We removed all silent and corrupted audio files. In the transcriptions, all numerical representations like numbers, dates, and times are replaced with their text equivalents according to the utterance. We note, however, that all the speakers read the numerical numbers in English and not in the respective native language, as is currently common in Zambia. We attribute this to the influence of the English language which is the national official language of communication and education. All the speakers are bilingual, fluent in both their native language and English. 

\noindent \textbf{Data splits} \ 
In Table~\ref{labelled-datasets}, we provide fixed splits for the labelled datasets. For Nyanja, whose transcriptions are mapped to the English captions of images in the Flickr30K~\cite{plummer2015flickr30k}, we provide two sets of splits. In the first set, we use the original Flickr30K~\cite{plummer2015flickr30k} splits to provide mappings for training (22 hrs), development (50 minutes), testing (50 minutes), and held out (50 minutes) sets. The \textit{held out} set is the number of samples that could not be aligned to either of the original splits. Unfortunately, this strategy does not guarantee the avoidance of speaker overlap among the splits. We, therefore, consider these to be our \textit{secondary splits}. In our \textit{primary splits}, we follow as much as possible the ratio 80\%:10\%:10\% while considering gender distribution and speaker overlap avoidance to split the datasets. The same strategy is adopted for Tonga and Lozi datasets, leading to splits without speaker overlap.

\subsection{Unlabelled datasets} \ 
\label{sec:unannotated-datasets}
In this section, we detail the creation process as well as present general statistics of the unannotated datasets. The unlabelled audio collections correspond to 357 hours of radio news and talk shows recordings for five languages; Bemba, Nyanja, Tonga, Lozi, and Lunda. We automatically segmented the audio data generating 168 hours of usable speech data for speech processing of lengths between 1 and 30 seconds. We provide the summary statistics for the resulting unannotated datasets in Table~\ref{tab:unlabeled-datasets}. All audio files are encoded in Waveform Audio File Format (WAVE) with a dual-track (stereo) and recording with a sample rate of 44.1kHz.

\begin{table}[th]
	\caption{Unlabelled radio broadcast styled audio files for five (5) languages.}
	\label{tab:unlabeled-datasets}
	\centering
	\begin{tabular}{lcccc}
		\toprule 
		\textbf{Languages} & \textbf{} & \textbf{No. of files} & \textbf{No. of hours} & \textbf{}  \\
		\midrule
		\multicolumn{5}{l}{\textit{Original:}}\\
		\cmidrule(l){1-2}
		Bemba &  & 533 & 162  \\
		Nyanja & & 26 & 25   \\
		Tonga & & 122 & 101  \\
		Lozi & & 37 & 30 \\
		Lunda & & 50 & 39 \\
		\midrule
		total & & 768 & 357   \\
		\midrule
		\multicolumn{5}{l}{\textit{Segmented:}}\\
		\cmidrule(l){1-2} 
		Bemba &  & 26,855 & 63  \\
		Nyanja & & 6,976 & 10   \\
		Tonga & & 38,012 & 60  \\
		Lozi & & 8,845 & 15 \\
		Lunda & & 13,424 & 20 \\
		\midrule
		total &  & 89,517 & 168  \\
		\bottomrule	
	\end{tabular}
\end{table}

\noindent \textbf{Data collection} \ 
We collected audio recordings of TV/radio news and talk show broadcast items from various radio stations and YouTube. This was achieved through a \textit{data sourcing sprint} event that we set up and invited students to participate in by collecting recordings of news items and talk show items in local languages from various TV and radio stations. 

\noindent \textbf{Segmentation} \ 
\label{sec:segmentation-process}
We segmented the audio files using segmentation and voice activity detection modules of \textit{pyannote tool}~\cite{Bredin2020,Bredin2021} in order to (i) produce smaller audio chunks compatible with present speech processing models and (ii) to automatically remove all non-speech events such as silence and music. We removed all the audio files with a duration of less than 1 second and greater than 30 seconds similar to the preprocessing of~\cite{zanon-boito-etal-2022-speech} for Tamasheq audio collections. The results of this step are presented in Table~\ref{tab:unlabeled-datasets}, ending up with 63 hours of Bemba, 10 hours of Nyanja, 60 hours of Tonga, and 15 hours of Lozi. 

\noindent \textbf{Resulting corpus} \ 
We make available both the original and segmented audio files to the research community. We release the unsegmented audio recordings to allow the research community to choose their own segmentation approach for their experiments.

\section{Experiments}
\label{sec:baseline-experiments}
For our baseline experiments, we build end-to-end (E2E) monolingual speech recognition models for each language as well as a multilingual model. We explore training models from scratch and leverage cross-lingual transfer learning by finetuning large-scale multilingual pretrained speech models on labelled datasets. We give details of our experiments below.

\subsection{Training from Scratch}
We follow the FairSEQ S2T toolkit~\cite{ott2019fairseq,wang2020fairseqs2t} implementation to train E2E ASR Transformer~\cite{Vaswani2017} models from scratch. We use the small Transformer~\cite{Vaswani2017} base architecture with 71 M parameters, \texttt{s2t\_transformer\_s}. We follow the data preprocessing of~\cite{wang2020fairseqs2t}, lowercasing and removing punctuation except for apostrophes for all transcriptions, and build 1K unigram subword vocabularies with 100\% character coverage using SentencePiece~\cite{kudo-richardson-2018-sentencepiece} without pre-tokenization. We train models for 500 epochs using the Adam optimiser~\cite{Kingma2015} with 10K warm-up steps on a single NVIDIA Tesla P100 GPU using the Google Colab+ platform. The models are optimised to minimise the \texttt{label\_smooth\_cross\_entropy} criterion function using a starting learning rate of \texttt{2e-3} with the label-smoothing coefficient of \texttt{0.1}. For inference, we use beam search with a beam size of 5 for decoding. We use the best checkpoints for evaluation. In Table~\ref{tab:baseline-results}, we report the models' performance on the test sets as measured by the word error rate (WER) metric. 

\subsection{Finetuning Multilingual Pretrained Models}
Cross-lingual transfer learning has been shown to be effective in improving speech recognition in low-resource languages~\cite{ babu22_interspeech,Baevski2020, Conneau2021}. We explore this strategy by finetuning XLS-R~\cite{babu22_interspeech}, a Wav2Vec2.0\cite{Baevski2020} based large-scale multilingual pretrained model on the labelled language datasets to obtain both monolingual and multilingual speech recognition models. We use the 300 Million parameter model (XLS-R-0.3B)~\cite{babu22_interspeech} for our experiments. We finetune the XLS-R(0.3B)~\cite{babu22_interspeech} on each individual language dataset (\#lang=1) to obtain monolingual models. For the multilingual model, we multilingually finetune the XLS-R(0.3B)~\cite{babu22_interspeech} on the labelled data of all four (4) languages (\#lang=4): Bemba, Nyanja, Tonga and Lozi.  All models are trained using the HuggingFace Transformer library~\cite{Wolf_Transformers_State-of-the-Art_Natural_2020} with Connectionist Temporal Classification~\cite{Graves2006}. We inherit default training configurations from the library except for batch size, learning rate and mask probabilities. All models are trained for 5  epochs with the learning rate of \texttt{3e-4} on a single NVIDIA Tesla P100 GPU using the Google Colab+ platform. We utilize gradient checkpoint and accumulation to optimize GPU memory usage and utilization. We also use early stopping based on dev set performance to avoid model overfitting. 

In addition, we also train 5-gram language models (LMs) using text from the training and development set from each language dataset using the KenLM~\cite{Heafield2011} toolkit to improve model performance. We use the python-based CTC beam search decoder library, \texttt{pyctcdecode}\footnote{\url{https://github.com/kensho-technologies/pyctcdecode}} to connect the LM to the finetuned models. For both sets of models, with and without a LM, we report the WER results on test sets in Table~\ref{tab:baseline-results}.

\section{Results}
\label{sec:results}
Table~\ref{tab:baseline-results} summarises the results of our baseline experiments. First, we note that the performance of the models trained from scratch is unacceptable for the lower-resource settings (toi and loz have WER around 90). By finetuning XLS-R, on the other hand, with basic training configurations, we obtain results that could be considered relatively acceptable, given the small size of our datasets. Incorporating the symbolic LM in the decoding process yields additional significant improvements for all languages except Bemba. In addition, we note also that multilingually finetuning the XLS-R on the four labelled language datasets does not yield better performance compared to monolingually trained models except for Nyanja (22.45) with the inclusion of a LM. 

The performances still depict the challenge of end-to-end speech recognition in a low-resource setting. There are several directions we can take to improve the results. For instance, we could explore self-training and pseudo-labelling strategies by leveraging the unlabelled audio collections presented in Section~\ref{sec:unannotated-datasets} which we were unable to do because of the lack of appropriate compute resources.

\begin{table}[th]
	\caption{Models relying on the pre-train and finetune paradigm outperform models trained from scratch. Only Nyanja (nya) benefits from multilingual finetuning, but including a language model leads to improvements for all languages. Showing test set results using word error rate (WER).}
	\label{tab:baseline-results}
	\centering
	\begin{tabular}{lc|cccc}
		\toprule 
		\textbf{Model}  & \textbf{\#lang} &  \textbf{bem} & \textbf{nya} & \textbf{toi} & \textbf{loz}  \\
		\midrule
		\multicolumn{4}{l}{from scratch: } \\
        \ \ S2T& & 40.13 & 43.20 & 89.36 & 91.11 \\
        \multicolumn{4}{l}{pretrain + finetune (monolingual):} \\
        \ \ XLS-R & 1 &34.50 & 30.18 & 35.65 & 46.64 \\
        \ \ XLS-R + LM & 1 &\textbf{34.28} & 27.28 & \textbf{28.09} & \textbf{34.59} \\
        \multicolumn{4}{l}{pretrain + finetune (multilingual):} \\
        \ \ XLS-R & 4 & 44.89 & 35.06 & 47.31 & 64.67 \\
        \ \ XLS-R + LM & 4 & 43.68 & \textbf{22.45} & 35.10 & 34.83 \\
		\bottomrule
	\end{tabular}
\end{table}

\section{Conclusion}
\label{sec:conclusion}
In this paper, we presented an open-source, multilingual speech corpus for the low-resourced native languages of Zambia. The corpus contains two collections of speech resources for Zambian languages; labelled and unlabelled. The labeled portion contains 80 hours of speech for four languages; Bemba (28 hrs), Nyanja (25 hrs), Tonga (22 hrs), and Lozi (6 hrs) while the unlabelled radio broadcast audio collection (357 hours) which has been segmented automatically to 168 hours for the same four languages plus  Lunda. We hope that by releasing the dataset we will encourage research and development of speech-based systems that support native languages of Zambia. 

In future work, we plan to explore methods that will take advantage of the unlabelled audio data presented in Section~\ref{sec:unannotated-datasets}.\footnote{We are currently unable to do so due to the lack of adequate computational resources at our institutions.} In addition, we aim to improve the datasets in both size and the number of languages. In the immediate future, our goal is to expand our speech data resources to all seven official languages of Zambia.


\section{Acknowledgements}
We would like to thank all the participants that made this work possible. We also would like to extend our appreciation to all the contributors of radio broadcast audio collections to this work. Antonios Anastasopoulos is generously supported by NSF-NEH grant BCS-2109578. 
\bibliographystyle{IEEEtran}
\bibliography{mybib}

\end{document}